\RequirePackage[l2tabu, orthodox]{nag}
\documentclass[11pt]{article}

\input{preamble/format}

\usepackage[usenames,dvipsnames]{xcolor}

\usepackage{lineno}
\usepackage{graphicx}
\usepackage[labelfont=bf]{caption}
\usepackage[format=hang]{subcaption}

\usepackage[algoruled,ruled,vlined]{algorithm2e}
\SetKwInput{KwControl}{Controls}

\usepackage{listings}
\usepackage{fancyvrb}
\fvset{fontsize=\normalsize}

\lstdefinestyle{mystyle}
{
    commentstyle=\color{OliveGreen},
    keywordstyle=\color{BurntOrange},
    numberstyle=\tiny\color{black!60},
    stringstyle=\color{darkblue},
    basicstyle=\ttfamily,
    breakatwhitespace=false,
    breaklines=true,
    captionpos=b,
    keepspaces=true,
    numbers=left,
    numbersep=5pt,
    showspaces=false,
    showstringspaces=false,
    showtabs=false,
    tabsize=2
}
\usepackage{enumitem}

\usepackage{natbib}
\usepackage[colorlinks,linktoc=all]{hyperref}
\usepackage[all]{hypcap}
\hypersetup{citecolor=MidnightBlue}
\hypersetup{urlcolor=MidnightBlue}
\hypersetup{linkcolor=black}

\usepackage[nameinlink]{cleveref}
\creflabelformat{equation}{#1#2#3}
\crefname{equation}{eq.}{eqs.}
\Crefname{equation}{Eq.}{Eqs.}
\Crefname{section}{Section}{Sections}

\usepackage
[acronym,nowarn,section,nogroupskip,nonumberlist]{glossaries}
\glsdisablehyper{}
\setglossarystyle{long3col}
\makeglossaries

\usepackage{blindtext}
\usepackage{framed}

\usepackage{booktabs}
\usepackage{multirow}
\usepackage{longtable}
\usepackage{etoolbox,siunitx}
\robustify\bfseries
\newcommand{\parhead}{\paragraph}

\usepackage[colorinlistoftodos,
prependcaption,
textsize=footnotesize,
backgroundcolor=blue!10,
linecolor=black!10,
bordercolor=yellow!20]{todonotes}

\newcommand{\dd}{\mathrm{d}}

\newcommand{\g}{\,\vert\,}
\newcommand{\s}{\,;\,}

\newcommand{\EE}[2]{\mathbb{E}_{#1}\left[#2\right]}

\newcommand{\kl}[1]{\mathrm{KL}\left(#1\right)}

\newcommand{\rmp}{\mathrm{p}}
\newcommand{\rmq}{\mathrm{q}}

\newcommand{\cL}{\mathcal{L}}

\newcommand{\mbx}{\mathbf{x}}

\newcommand{\mbz}{\mathbf{z}}

\newtheorem{theorem}{Theorem}
\newtheorem{proposition}{Proposition}
\newtheorem{definition}{Definition}

\newtheorem{corollary}{Corollary}

\usepackage{siunitx}
\usepackage{titlesec}

\title{Simulation-Based Empirical Bayes}
\author{
  Xinwei Shen\\
  Department of Statistics\\
  University of Washington
  \and
  Diana Cai\\
  Department of Computer Science\\
  Cornell University
  \and
  Cheng Zhang\\
  School of Mathematical Sciences and Center for Statistical Science\\
  Peking University
  \and
  David M. Blei\\
  Departments of Computer Science and Statistics\\
  Columbia University
}

\begin{document}
\date{}
\maketitle

\begin{abstract}
  Empirical Bayes (EB) performs simultaneous inference across many related latent variables.  Classical EB assumes that the likelihood $\rmp(x \g z)$ is tractable.  In many scientific applications, however, the likelihood is available only through a simulator.  This paper develops EB for such implicit likelihoods.  We introduce simulation-based empirical Bayes (SBEB), which connects nonparametric EB to simulation-based inference (SBI).  SBEB computes EB estimates without an explicit density by using the observed data, simulator samples, and an amortized inference network. SBEB iteratively refines the fitted EB prior toward the population prior. With several scientific simulators and real-world data, we demonstrate that SBEB improves accuracy over SBI with a fixed prior.
\end{abstract}


\section{Introduction}

\begin{figure}[t]
  \begin{center}
    \includegraphics[width=0.9\textwidth]{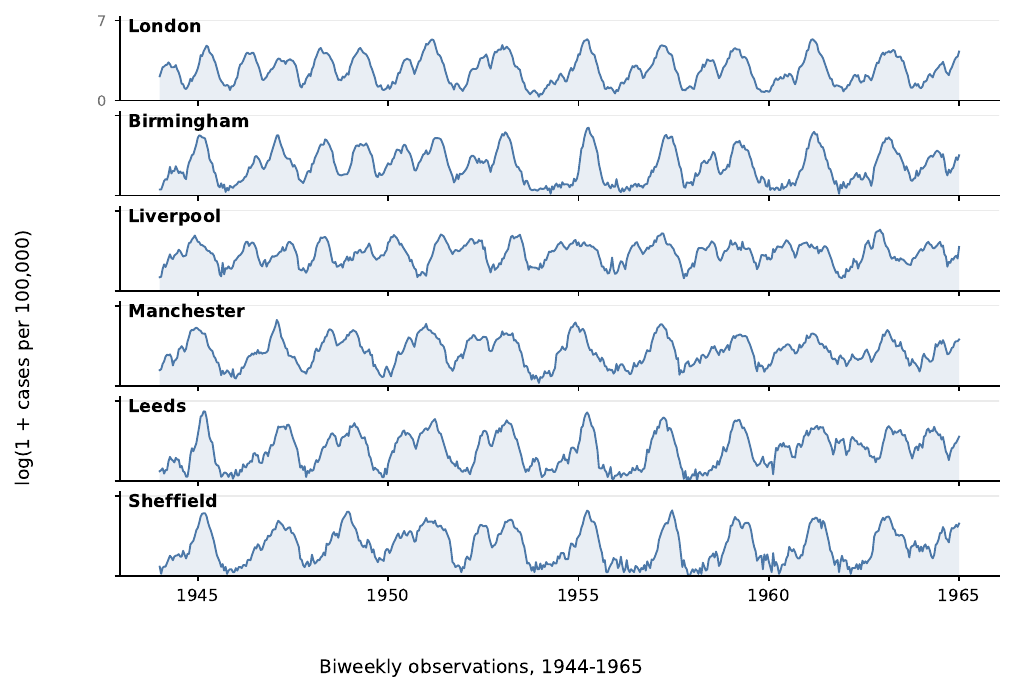}
  \end{center}
  \caption{Biweekly measles reports from six UK cities.  Each panel is
    one city.  The full dataset contains 40 cities observed from 1944
    to 1965 \citep{Dalziel:2016}.}
  \label{fig:measles-trajectories}
\end{figure}

This paper is about simultaneous inference with simulation-based likelihoods and empirical Bayes. As an example, \Cref{fig:measles-trajectories} shows biweekly measles reports from six UK cities, drawn from a dataset of 40 cities.  The trajectories show recurrent outbreaks, changes in amplitude, and city-to-city variation.

To explain data like these, epidemiologists build mechanistic simulators \citep{FinkenstadtGrenfell:2000,Bjornstad:2002}.  In such a simulator, the input is a vector of epidemic parameters for transmission, seasonality, and reporting.  The output is a time series of reported cases.  These epidemic simulators encode theories about how infections spread and are reported.

Given data like those in \Cref{fig:measles-trajectories}, the statistical goal is to infer the epidemic parameters for each city.  Formally, let $x_i$ denote the observed time series for city $i$, and let $z_i$ denote its epidemic parameters.  A Bayesian analysis targets
\begin{align*}
  \rmp(z_i \g x_i) \propto \rmp(z_i)\rmp(x_i \g z_i),
\end{align*}
where $\rmp(z)$ is a prior over simulation parameters and
$\rmp(x \g z)$ is the conditional likelihood of a trajectory induced
by the simulator.

This problem has two notable features.  First, the likelihood is defined by simulation.  For a proposed parameter $z_i$, the simulator can generate a time series $x_i$, but it cannot necessarily evaluate $\rmp(x_i \g z_i)$.  Second, the problem is simultaneous.  We have many related cities and many related inference problems.  Their epidemic parameters differ, but they come from a common population.

These two features of the problem suggest two statistical ideas. The first idea is \textit{simulation-based inference} (SBI) \citep{Cranmer:2020,Deistler:2025}.  SBI uses simulator draws to approximate posterior inference when the likelihood cannot be evaluated.  It has become a standard tool for scientific models where simulation is available but likelihood calculation is not.  The SBI workflow begins by choosing a prior, drawing $z_b \sim \rmp(z)$, simulating $x_b \sim \rmp(x \g z_b)$, and fitting an amortized approximation $\rmq(z \s x)$ on these simulated pairs.  In SBI, the prior plays its usual role in the posterior, but it is also used to draw the parameters used for simulation.

The second idea is \textit{empirical Bayes} (EB) \citep{Robbins:1956,Stein:1956,Efron:1973,Efron:2014,Efron:2019}.  EB uses the collection of observations to learn the prior for the latent parameters.  This helps simultaneous inference in two ways.  From a Bayesian view, the learned prior captures the population that generated the units.  From a frequentist view, the learned prior shrinks noisy unit-level estimates toward the population and can improve repeated-sampling risk.

In this paper, we combine these two ideas to develop \textit{simulation-based empirical Bayes} (SBEB), a method that uses SBI to perform simultaneous inference with an EB prior.  SBEB refines an approximate nonparametric EB prior and samples from it inside an SBI loop.  The result is EB analysis for intractable, simulation-based likelihoods.

SBEB extends both SBI and EB\@.  In standard SBI, the prior is fixed before simulation and training \citep{Cranmer:2020,Deistler:2025}.  SBEB instead learns a flexible prior from the data, while it performs simulation-based inference.  Classical EB methods usually rely on simple, explicit likelihoods, such as Gaussian normal-means models \citep{Stein:1956,Efron:1973} or Poisson models \citep{Robbins:1956,Buhlmann:1967}.  SBEB extends empirical Bayes to implicit likelihoods based on complex simulators.

After developing SBEB, we study it theoretically and
empirically. Theoretically, we study an oracle version and prove that
it converges to the population prior, i.e., the prior whose induced
marginal distribution of observations matches the population.  Empirically, we first study SBEB in controlled simulators, where the true latent variables are known.  We then study two real datasets: discrete choice models for product choices by people~\citep{Allenby:2014:QME,Allenby:2014:JLE} and epidemic simulators for measles time series by cities~\citep{Dalziel:2016}.  Across these studies, SBEB improves on classical SBI with a fixed prior.


\section{Simulation-based empirical Bayes}

We develop the method in three steps.  First, we review empirical Bayes (EB) for simultaneous inference and define the population prior.  Second, we review simulation-based inference (SBI), which fits an amortized posterior approximation from simulator draws.  Finally, we develop simulation-based empirical Bayes (SBEB). It iteratively uses simulation-based inference to approximate posterior distributions, and then uses those posteriors to refine an empirical Bayes prior.

\subsection{Empirical Bayes and the population prior}

Consider a model with local latent variables,
\begin{align}
  \label{eq:local_model}
  \rmp(z, x) = \rmp(z) \rmp(x \g z).
\end{align}

We observe a dataset $\mbx = x_{1:n}$. Our goal is
\textit{simultaneous inference} of $\mbz = z_{1:n}$,
\begin{align}
  \label{eq:simul_inf}
  \rmp(\mbz \g \mbx) = \prod_{i=1}^{n} \rmp(z_i \g x_i).
\end{align}
This problem involves $n$ posterior inferences.  The prior $\rmp(z)$ ties these inferences together. The empirical Bayes question is: what should the prior $\rmp(z)$ be?

Suppose our data came from a true population distribution,
$x_i \overset{\rm i.i.d.}\sim \rmp^\star(x)$ .  The \textit{empirical Bayes criterion}
says that the model marginal should match the true population
distribution,
\begin{align}
  \label{eq:eb_criterion}
  \rmp^\star(x) = \rmp(x),
\end{align}
where the marginal is
\begin{align}
  \label{eq:marginal}
  \rmp(x) = \int \rmp(z) \, \rmp(x \g z) \, \dd z.
\end{align}
The goal of empirical Bayes is to set the prior $\rmp(z)$ so that the
EB criterion holds~\citep{Robbins:1956,Efron:2019}.


A prior that satisfies the EB criterion can be written as an expectation
over the posterior,
\begin{align}
  \label{eq:pop_prior}
  \rmp(z) = \EE{\rmp^\star(x)}{\rmp(z \g x)}.
\end{align}
We call this prior the \textit{population prior}.  The equation is an
identity: the prior $\rmp(z)$ appears on both sides because the
posterior $\rmp(z \g x)$ is computed under that same prior.  Related
posterior-averaging forms appear in nonparametric empirical Bayes and
amortized variational inference
\citep{Laird:1978,McInerney:2015,Tomczak:2017}.  We prove
\Cref{eq:pop_prior} in \Cref{app:proofs}.

The identity in \Cref{eq:pop_prior} does not define the prior directly; it gives a fixed point.  But it suggests an algorithm: if we can approximate the posteriors $\rmp(z \g x_i)$ for the observed units, then we can approximate the population prior by averaging those posteriors.  We use this idea to develop SBEB in \Cref{sec:sbeb}.

\subsection{Fixed-prior simulation-based inference}

Suppose we can simulate from $\rmp(x \g z)$ but cannot evaluate its
density.  For now, suppose the prior $\rmp(z)$ is fixed.  (We will
return to EB later.)  This subsection explains fixed-prior
simulation-based inference in three steps: the amortized family, the
training objective, and the simulator-based Monte Carlo approximation.

Given a dataset $\mbx$, how can we perform simultaneous inference?
One strategy is \textit{simulation-based inference}
(SBI)~\citep{Papamakarios:2019,Cranmer:2020}.
A common SBI approach, neural posterior estimation, uses simulated data
to fit an \textit{inference network} $\rmq(z \s x, \varphi)$ that
approximates the posterior for any observation $x$.
For the observed units, we use $\rmq(z \s x_i, \varphi)$ as an
approximation to $\rmp(z \g x_i)$.

The inference network is amortized: the same parameter vector
$\varphi$ is used for all observations.  The network takes an
observation $x$ as input and returns a distribution over $z$.  For
example, the output might be the mean and covariance of a Gaussian
distribution, or the parameters of a more flexible conditional density
model.  Neural networks and normalizing flows are common choices
\citep{Papamakarios:2016,Greenberg:2019}.

The target is the exact posterior $\rmp(z \g x)$, averaged over the
kinds of observations generated by the model.
We fit the inference parameters $\varphi$ by optimizing the SBI objective,
\begin{align}\label{eq:sbi_obj}
  \cL(\varphi) = \EE{\rmp(x)}{\kl{\rmp(z \g x) \, \Vert \, \rmq(z \s x, \varphi)}}.
\end{align}
It is the expected KL divergence between the exact and approximate posteriors, where the expectation is with respect to the marginal distribution of the observation.

The key idea behind SBI is that the same joint distribution can be
written in two ways,
$\rmp(x)\rmp(z \g x) = \rmp(z)\rmp(x \g z)$.  We cannot sample from
$\rmp(x)\rmp(z \g x)$ directly, because that would require posterior
samples.  But with a fixed prior and a simulator, we can sample from
$\rmp(z)\rmp(x \g z)$.  This lets us rewrite the objective,
\begin{align}
  \label{eq:sbi_1}
  \cL(\varphi)
  &=
    \EE{\rmp(x)}{\EE{\rmp(z \g x)}{\log \rmp(z \g x) - \log \rmq(z \s
    x, \varphi)}}
  \\
  \label{eq:sbi_2}
  &= \EE{\rmp(z, x)}{\log \rmp(z \g x) - \log \rmq(z \s x, \varphi)}.
\end{align}
\Cref{eq:sbi_1} uses the definition of the KL divergence; \Cref{eq:sbi_2} uses that $\rmp(z, x) = \rmp(x) \, \rmp(z \g x)$. We have now written the objective in terms of an expectation with respect to the joint.

The first term in \Cref{eq:sbi_2} does not depend on $\varphi$. We minimize the second term with a Monte Carlo approximation,
\begin{align*}
  \arg \min_{\varphi} \cL(\varphi) &= \arg \min_{\varphi} \EE{\rmp(z, x)}{- \log \rmq(z \s x, \varphi)} \\
                                   &\approx
                                     \arg \min_{\varphi} \frac{1}{B} \sum_{b} - \log \rmq(z_b \s x_b, \varphi)
                                     \quad \quad z_b, x_b \sim \rmp(z) \, \rmp(x \g z).
\end{align*}
Thus fixed-prior SBI reduces posterior approximation to supervised
learning on simulated pairs $(z_b,x_b)$.

\Cref{alg:sbi} summarizes fixed-prior SBI. Its output is a fitted inference network $\hat{\varphi}$. For each observed unit $x_i$, we approximate the posterior with $\rmq(z \s x_i, \hat{\varphi})$.

\begin{algorithm}[t]
  \DontPrintSemicolon

\BlankLine

\KwIn{
  \begin{itemize}[noitemsep]
  \item $\mbx = \{x_i\}_{i=1}^{n}$: a dataset
  \item $\rmp(z)$: a prior
  \item $\rmp(x \g z)$: a simulator
  \item $\rmq(z \s x, \varphi)$: an inference network
  \end{itemize}}

\BlankLine

\KwControl{
  \begin{itemize}[nosep]
  \item $B$: the number of Monte Carlo samples
  \end{itemize}
}

\BlankLine

\For{$b$ in $1 \ldots B$}
{
  sample $z_b \sim \rmp(z)$
  
  simulate $x_b \sim \rmp(x \g z_b)$
}

set
$\hat{\varphi}
\leftarrow
\arg \min_{\varphi}
\frac{1}{B} \sum_{b=1}^{B} - \log \rmq(z_b \s x_b, \varphi)$

\Return{$\{\rmq(z \s x_i, \hat{\varphi})\}_{i=1}^n$}


  \caption{Simulation-based inference}
  \label{alg:sbi}
\end{algorithm}

\subsection{Simulation-based empirical Bayes}\label{sec:sbeb}

We now return to the empirical Bayes problem.  The prior should be the population prior in \Cref{eq:pop_prior}, but the likelihood is only available through simulation.  We cannot use classical EB methods, which require evaluating $\rmp(x \g z)$, and we cannot use traditional SBI, which requires a pre-specified prior.  \textit{Simulation-based empirical Bayes} (SBEB) combines the two ideas: it learns an approximate population prior and uses it inside an SBI loop.

SBEB works in rounds.  At round $t$, the current inference network
gives an approximate posterior for each observed unit, $\rmq(z \s
x_i,\hat{\varphi}_{t-1})$.  We average these approximate posteriors
over the observed units:
\begin{align}
  \label{eq:approx_pop_prior}
  \hat{\rmp}_t(z)
  =
  \EE{\hat{\rmp}_n(x)}{\rmq(z \s x, \hat{\varphi}_{t-1})}
  =
  \frac{1}{n}\sum_{i=1}^n \rmq(z \s x_i, \hat{\varphi}_{t-1}).
\end{align}
This is the sample analogue of the population-prior identity in \Cref{eq:pop_prior}, where $\hat{\rmp}_n(x)$ denotes the empirical distribution of the observed data.

The key point is that we can sample from $\hat{\rmp}_t(z)$ even though
we do not have a closed-form formula for its density.  To draw from it,
sample an observed unit $x_i$ uniformly from the dataset, and then draw
$z \sim \rmq(z \s x_i,\hat{\varphi}_{t-1})$.  Given this draw of $z$,
we simulate a new observation $x \sim \rmp(x \g z)$.  These simulated
pairs train the next inference network.

\Cref{alg:sbi_eb} summarizes SBEB\@.  It starts from an initial prior $\rmp_0(z)$ and fits a fixed-prior SBI approximation. It then iteratively draws parameters from the approximate EB prior, simulates new observations, and refits the inference network.  It returns posterior approximations under the learned population prior, $\{\rmq(z \s x_i, \hat{\varphi}_T)\}_{i=1}^n$.

\begin{algorithm}[t]
  \DontPrintSemicolon

\BlankLine

\KwIn{
  \begin{itemize}[noitemsep]
  \item $\mbx = \{x_i\}_{i=1}^{n}$: a dataset
  \item $\rmp_0(z)$: an initial prior
  \item $\rmp(x \g z)$: a simulator
  \item $\rmq(z \s x, \varphi)$: an inference network
  \end{itemize}}

\BlankLine

\KwControl{
  \begin{itemize}[nosep]
  \item $B$: the number of Monte Carlo samples
  \item $T$: the number of rounds
  \end{itemize}
}

\BlankLine

fit $\hat{\varphi}_0$ by fixed-prior SBI using $\rmp_0(z)$

\For{$t$ in $1 \ldots T$}
{
  \For{$b$ in $1 \ldots B$}
  {
    sample $\tilde{x}_b$ from the dataset $\mbx$
    
    sample $z_b \sim \rmq(z \s \tilde{x}_b, \hat{\varphi}_{t-1})$

    simulate $x_b \sim \rmp(x \g z_b)$
  }
  
  set
  $\hat{\varphi}_t
  \leftarrow
  \arg \min_{\varphi}
  \frac{1}{B} \sum_{b=1}^{B} - \log \rmq(z_b \s x_b, \varphi)$
}

\Return{$\{\rmq(z \s x_i, \hat{\varphi}_T)\}_{i=1}^n$}


  \caption{Simulation-based empirical Bayes}
  \label{alg:sbi_eb}
\end{algorithm}


\section{Convergence of the oracle population-prior update}

This section studies the idealized SBEB update.  We assume that the
fitted inference network
can represent the exact posterior under the current
prior, and that population expectations are available.  This separates
the statistical idea from the numerical approximations in the
algorithm.  Under these assumptions, SBEB iterates a map on priors and
moves toward the population prior.
The proofs are provided in \Cref{app:proofs}.

Let $\rmp_t(z)$ be the prior at iteration $t$, and let
$\rmp_t(x)=\int \rmp_t(z)\rmp(x \g z)\,\dd z$ be its induced model
marginal.  The posterior under this prior is
\begin{align*}
  \rmp_t(z \g x)
  =
  \frac{\rmp_t(z)\rmp(x \g z)}{\rmp_t(x)}.
\end{align*}
The population-prior identity
in \Cref{eq:pop_prior}
suggests the update
\begin{align}
  \label{eq:oracle_prior_update}
  \rmp_{t+1}(z)
  =
  \EE{\rmp^\star(x)}{\rmp_t(z \g x)}.
\end{align}
This is the population version of the SBEB update in
\Cref{eq:approx_pop_prior}.  The algorithm replaces the posterior
$\rmp_t(z \g x)$ with an amortized approximation and replaces the
population expectation with an average over observed units.

The update in \Cref{eq:oracle_prior_update} is also an expectation-maximization (EM) update.
To see this, we define the objective
\begin{align}
\label{eq:lowerbound}
\ell(\rmp(z), \rmq(z \s x, \varphi))
=
\EE{\rmp^\star(x)}
{\EE{\rmq(z \s x, \varphi)}
{\log \frac{\rmp(z)\rmp(x \g z)}
{\rmq(z \s x, \varphi)}}}.
\end{align}
For a fixed prior $\rmp_t(z)$, maximizing $\ell$ over $\rmq$ gives the
posterior $\rmp_t(z \g x)$.  For a fixed posterior approximation,
maximizing $\ell$ over the prior gives the average posterior in
\Cref{eq:oracle_prior_update}.  The idealized SBEB update is
coordinate ascent on this objective.

We analyze the idealized SBEB update in three stages.
First, we show that at every iteration, the update improves the fit of the induced marginal.
Next, we show that the prior KL divergence
decreases by at least the current marginal KL divergence;
this result then implies convergence of the sequence of marginals to the population marginal.
Finally, we introduce a prior identifiability condition
under which marginal convergence leads to convergence of the sequence of priors.

The first result says that the oracle update improves the model
marginal distribution of the data.
\begin{proposition}[Monotonicity]\label{prop:monotonicity}
    For the oracle update in \Cref{eq:oracle_prior_update}, it holds that
\begin{align*}
\EE{\rmp^\star(x)}{\log \rmp_t(x)}\, \leq \, \EE{\rmp^\star(x)}{\log \rmp_{t+1}(x)} ,
\end{align*}
Equivalently, the KL divergence from the population distribution to the
model marginal decreases,
\begin{align}
\kl{\rmp^\star(x) \, \Vert \, \rmp_{t+1}(x)} \,\leq \, \kl{\rmp^\star(x) \, \Vert \, \rmp_{t}(x)}.
\end{align}
\end{proposition}

Now let $\rmp^\star(z)$ be a population prior as defined in
\Cref{eq:pop_prior}, and let $\rmp^\star(x)$ be its induced marginal.
The next result says that each step also moves the
prior toward the population prior.

\begin{proposition}[Prior descent]\label{prop:dlemma}
For any $t\geq 0$,
\begin{align*}
\kl{\rmp^\star(z) \, \Vert \, \rmp_{t}(z)} - \kl{\rmp^\star(z) \, \Vert \, \rmp_{t+1}(z)} \geq \kl{\rmp^\star(x) \, \Vert \, \rmp_{t}(x)}.
\end{align*}
\end{proposition}

Together, the two propositions imply convergence of the model marginal.
Let $\rmp_0(z)$ be the initial prior.

\begin{corollary}[Marginal convergence]\label{cor:marginal-convergence}
Suppose that $\kl{\rmp^\star(z) \, \Vert \, \rmp_0(z)} < \infty$.
Then
\begin{align*}
\kl{\rmp^\star(x) \, \Vert \, \rmp_t(x)} \to 0,
\quad \textrm{as }\; t \to \infty .
\end{align*}
\end{corollary}

The convergence theorem for the prior needs one property of the
simulator.  If two priors are far apart, then their induced marginal
distributions of the data must also be far apart.  Otherwise the data
cannot distinguish the priors.

\begin{definition}[Prior identifiability]
\label{def:prior}
Let $\rmp_a(z)$ and $\rmp_b(z)$ be any two prior distributions. The corresponding marginal likelihoods after simulation are $\rmp_a(x) = \int \rmp_a(z)\rmp(x\g z)\, \dd z$ and $\rmp_b(x) = \int \rmp_b(z)\rmp(x\g z)\, \dd z$. We say that the simulator is prior-identifiable if
\begin{equation*}
\rmp_a(x) = \rmp_b(x), \; \forall \, x \quad \Rightarrow \quad \rmp_a(z) = \rmp_b(z), \; \forall \, z.
\end{equation*}
We say it is $\lambda$-strongly prior-identifiable, under KL
divergence, if
\begin{equation*}
\kl{\rmp_a(z) \, \Vert \, \rmp_b(z)} \, \leq \, \tfrac{1}{\lambda} \kl{\rmp_a(x) \, \Vert \, \rmp_b(x)}.
\end{equation*}
\end{definition}

{\bf Remark}. The simulator maps priors on $z$ to marginals on $x$
through the Markov kernel $\rmp(x \g z)$.  By the data processing
inequality for KL divergence~\citep{CoverThomas:2006},
\begin{align*}
\kl{\rmp_a(x) \, \Vert \, \rmp_b(x)}
\leq
\kl{\rmp_a(z) \, \Vert \, \rmp_b(z)}.
\end{align*}
Thus $\lambda$-strong prior-identifiability is an inverse
data-processing condition, so it can hold only with $\lambda \leq 1$.
This condition may hold only locally, when the two priors are close:
for some $R>0$, $\kl{\rmp_a(z) \, \Vert \, \rmp_b(z)} \leq R$.  The
convergence result below still holds in this local form when the
initial prior is close to the population prior.

Ordinary prior-identifiability gives a qualitative consequence of
\Cref{cor:marginal-convergence}.  By Pinsker's inequality, the marginal
KL convergence implies convergence in total variation, and hence weak
convergence, of $\rmp_t(x)$ to $\rmp^\star(x)$.  If the priors are tight
and the simulator maps weakly convergent priors to weakly
convergent marginals, then every weak limit point of $\rmp_t(z)$ must
induce $\rmp^\star(x)$; by prior-identifiability, that limit point is
$\rmp^\star(z)$.  Thus the priors converge weakly, though not under a
specified metric or at a specified rate.

Stronger identifiability gives more: it quantifies how marginal
discrepancy controls prior discrepancy.  The next theorem uses this
condition to obtain geometric convergence of the prior under the oracle
update.

\begin{theorem}\label{thm:convergence}
Suppose that the simulator is $\lambda$-strongly prior-identifiable for some $\lambda \in (0,1]$
and that $\kl{\rmp^\star(z) \, \Vert \, \rmp_0(z)} < \infty$.
Then
\begin{align*}
\kl{\rmp^\star(z) \, \Vert \, \rmp_t(z)} \, &\leq \, (1-\lambda)^t \cdot \kl{\rmp^\star(z) \, \Vert \, \rmp_0(z)},\\
\kl{\rmp^\star(x) \, \Vert \, \rmp_t(x)} \, &\leq \, (1-\lambda)^t \cdot \kl{\rmp^\star(z) \, \Vert \, \rmp_0(z)}.
\end{align*}
\end{theorem}

The constant $\lambda$ controls the contraction in this theorem: larger
$\lambda$ gives
faster contraction, while smaller $\lambda$
corresponds to a simulator that makes different priors harder to differentiate.
In sum, this theorem describes the oracle version of SBEB\@.  When the
simulator identifies the prior and the posterior approximation is
exact, the population-prior update moves geometrically toward the
population prior.  The SBEB algorithm approximates this update with a
finite dataset and amortized posterior family.


\section{Related work}

\parhead{Empirical Bayes.} Empirical Bayes (EB) estimates prior structure from data by matching model and population behavior. Classical EB ideas go back to \citet{Robbins:1956}; for a modern perspective, see \citet{Efron:2019}.  The empirical Bayes criterion (\Cref{eq:eb_criterion}) follows this line: set the prior so that the model marginal equals the population distribution.

\parhead{Amortized variational inference and the population prior.}
Amortized inference fits a shared inference function across observations
or datasets.  Several papers work with the population prior, sometimes
under different names.  The population posterior of
\citet{McInerney:2015} averages posterior distributions across data
streams.  The amortized population posterior of \citet{Goyal:2017}
learns inference across a population of datasets.  The VampPrior of
\citet{Tomczak:2017} represents a prior as a mixture of variational
posteriors evaluated at learned pseudo-observations.  These methods show
that posterior averages can define useful priors.

\parhead{Simulation-based inference.}  Simulation-based inference (SBI)
uses simulators for models whose likelihoods cannot be evaluated. Early
approaches include approximate Bayesian computation
\citep{Beaumont:2002,Sisson:2018}.  Neural SBI methods train
conditional density models from simulated pairs, including neural
posterior estimation \citep{Papamakarios:2016,Greenberg:2019},
neural likelihood estimation \citep{Papamakarios:2019}, and related
methods \citep{Wildberger:2023,Hikida:2025}.  Broader surveys and
benchmarks appear in
\citet{Cranmer:2020,Lueckmann:2021,TejeroCantero:2020}.

Sequential SBI adapts the simulation distribution over rounds
\citep{Papamakarios:2016,Lueckmann:2017,Greenberg:2019}.
It focuses simulation on the
observed data, so that the fitted posterior approximation is accurate
where it will be used.  SBEB also changes the simulation distribution
over rounds and also focuses simulation on the observed units.  But it
does so while fitting an empirical Bayes prior.  This gives a simpler
algorithm in one sense: SBEB samples from the current prior estimate and
does not need to reweight samples to correct back to a fixed prior.

\parhead{Likelihood-free source distribution estimation.} Some SBI
papers call the unknown prior the \textit{source distribution}.  In this
language, source distribution estimation is close to empirical Bayes.
Neural empirical Bayes~\citep{Vandegar:2021} first learns a neural
surrogate likelihood from simulated pairs.  It then fits a
parameterized generative model for the source distribution by optimizing
an estimate of the marginal likelihood.  Sourcerer~\citep{Vetter:2024}
takes a different route: it is sample-based, targets a maximum-entropy
source distribution, and uses sliced-Wasserstein distance to compare
observed data to simulations.

SBEB differs from both approaches.  It does not first learn a surrogate
likelihood, and it does not solve a separate source-estimation problem.
Instead, it keeps prior learning and posterior approximation in one
loop.  Each round samples from the current average posterior, simulates
new data, and refits the amortized posterior approximation.  The fixed
point is the population-prior identity in \Cref{eq:pop_prior}.


\section{Empirical studies}
\label{sec:empirical-studies}

We evaluate simulation-based empirical Bayes (SBEB) in three settings.  The first setting is a suite of controlled simulation studies, where the latent variables are known and we can directly measure how well we recover them.  The second and third settings are real-data studies: heterogeneous product choice and childhood measles dynamics.  In the real-data studies, the latent variables are not observed, so we evaluate by holding out data from each unit and asking how well the inferred unit-specific parameters predict the held-out observations.

Across studies, a unit $i$ has observed data $x_i$ and a latent parameter $z_i$.  Fixed-prior SBI trains the amortized posterior from a fixed prior (\Cref{alg:sbi}).  SBEB starts from the same prior and then uses the collection of observed units to estimate an empirical Bayes prior while training the amortized posterior (\Cref{alg:sbi_eb}).  The real-data comparisons also include a per-unit method that does not share information across units.

\subsection{Simulation studies}
\label{subsec:simulation-studies}

We first study SBEB in simulation, where we know the latent variables that produced the data.  Each unit is generated by first drawing a latent parameter $z_i$ from a true prior and then drawing an observation from a stochastic simulator, $x_i \sim \rmp(x \g z_i)$.

We study five simulators:
\begin{itemize}
\item \textit{Linear Gaussian:} a conjugate Gaussian model with unit-specific
  means;
\item \textit{Nonlinear static:} a nonlinear regression model with stochastic
  observations;
\item \textit{Oscillator:} a stochastic observation of a standard dynamical
  system~\citep{Strogatz:2015};
\item \textit{Predator-prey:} a Lotka-Volterra population model with noisy
  observations~\citep{Lotka:1925,Volterra:1926};
\item \textit{Evolutionary:} a Wright-Fisher population-genetic simulator with
  drift and selection~\citep{Fisher:1930,Wright:1931}.
\end{itemize}
\Cref{app:simulation-details} gives the mathematical definition of each
simulator.

For each simulator, we compare fixed-prior SBI and SBEB\@. We also report a true-prior oracle.  The oracle trains SBI from the true prior; it is unavailable in real applications, but it shows how well the amortized family can do when the prior is correct.  All methods use the same diagonal-Gaussian amortized posterior: a two-layer perceptron maps the simulator output $x_i$ to a mean and log standard deviation for Gaussian $\rmq(z_i \s x_i, \varphi)$.

Because the data are simulated, we directly evaluate the estimated $z_i$.  Let
\begin{align*}
  \hat z_i =
  \EE{\rmq(z_i \s x_i, \hat{\varphi})}{z_i}
\end{align*}
be the posterior mean, and let $d_z$ be the dimension of $z_i$.  We report mean
squared error between the estimate $\hat{z}$ and the true $z$,
\begin{align*}
  \mathrm{mse}
  =
  \frac{1}{n}\sum_{i=1}^n \frac{1}{d_z}
  \left\| \hat z_i - z_i \right\|_2^2.
\end{align*}
Lower values are better.  Each run uses $n=500$ generated evaluation units with
known latent variables; each SBI training round uses 400 simulator-generated
training pairs.

\begin{table}[t]
\centering
\caption{Simulation studies.  Mean squared error of posterior means; lower is better. The oracle uses the true prior and is unavailable in real-data applications.}
\label{tab:simulation-mse}
\vspace{1ex}
\begin{tabular}{lccc}
\toprule
Simulator & Fixed prior & SBEB & True prior \\
\midrule
Linear Gaussian & 0.410 & 0.034 & 0.033 \\
Nonlinear static & 2.499 & 1.181 & 0.267 \\
Oscillator & 0.057 & 0.009 & 0.003 \\
Predator-prey & 0.569 & 0.371 & 0.006 \\
Evolutionary & 4.641 & 2.400 & 0.687 \\
\bottomrule
\end{tabular}
\end{table}

\Cref{tab:simulation-mse} shows the results. We see the same pattern across the five simulators.  SBEB has lower error than fixed-prior SBI\@.  In the linear Gaussian and oscillator studies, SBEB nearly matches the true-prior oracle.  In the nonlinear static, predator-prey, and evolutionary studies, the oracle remains better, but SBEB still reduces the error of fixed-prior SBI.

\subsection{Heterogeneous choice}
\label{subsec:choice-study}

We now turn to real-data studies.  Discrete choice models are a standard tool
in econometrics \citep{McFadden:1974,Train:2009}.  A discrete choice model
relates the features of a menu of alternatives to the alternative a person
chooses.  Our choice study uses the digital camera conjoint survey distributed
with the \texttt{bayesm} package~\citep{Rossi:2025:bayesm}.  The data are
described in \citet{Allenby:2014:QME,Allenby:2014:JLE}.

In this dataset, each unit is a person who faces 16 choice tasks.  In each
task, the person sees a menu of digital cameras, together with an outside
option, and chooses one alternative.  The camera alternatives have observed
attributes such as brand, product features, and price.  The goal is to infer a
person-specific preference vector from the first few choices and then predict
that person's held-out choices.

Let $x_{it}$ be the alternative chosen by person $i$ in task $t$, for
$t=1,\ldots,16$.  Let $a_{itj}$ be the feature vector for alternative $j$ in
task $t$, including features such as brand and price.  The latent parameter
$z_i$ is the person's vector of utility coefficients.  The simulator is the
multinomial logit choice model
\citep{McFadden:1974},
\begin{align*}
  x_{it} \sim \mathrm{Categorical}\{\pi_{it1}(z_i),\ldots,\pi_{itJ}(z_i)\},
  \qquad
  \pi_{itj}(z_i)
  =
  \frac{\exp(a_{itj}^{\mathsf T} z_i)}
       {\sum_{\ell=1}^J \exp(a_{it\ell}^{\mathsf T} z_i)}.
\end{align*}
This model has a random-utility interpretation.  In task $t$, alternative $j$
has utility $a_{itj}^{\mathsf T}z_i + \epsilon_{itj}$, where the errors
$\epsilon_{itj}$ are independent type-I extreme-value random variables.  The
person chooses the alternative with the largest utility.

For a training length $R$, the inference network receives a fixed-length
encoding $s_i^{(R)}$ of the menus $a_{i}$ and the observed choices
$x_{i,1:R}$.  The encoding includes the menu attributes for all 16 tasks.  For
tasks after $R$, the choice entry is masked.  The network maps $s_i^{(R)}$ to
a diagonal Gaussian approximation $\rmq(z_i \s s_i^{(R)}, \varphi)$.

We compare fixed-prior SBI, SBEB, and individual maximum likelihood fits.

The fixed prior is
\begin{align*}
  z_i \sim \mathcal{N}(0,\mathrm{diag}(\tau^2)),
\end{align*}
on the unconstrained utility coefficients, with $\tau=2$ for all coordinates
except the unconstrained price coefficient, where $\tau=1$.  The price
coefficient is transformed to be negative before simulation.  Fixed-prior SBI
shares an amortized inference network across simulated respondents, but it does
not learn a prior from the observed respondents.  SBEB starts from the same
broad prior and learns an empirical Bayes prior, following \Cref{alg:sbi_eb}.
The individual fit estimates one preference vector per person using only that
person's observed training tasks, with a weak Gaussian penalty whose standard
deviation is 5.

We train on the first $R$ choices, for different values of $R$, and always
evaluate on choices 13--16.  For a method with posterior samples
$z_i^{(1)},\ldots,z_i^{(S)}$, the held-out predictive probability is the
posterior average
\begin{align*}
  \hat\pi_{itj}
  =
  \frac{1}{S}\sum_{s=1}^S \pi_{itj}(z_i^{(s)}).
\end{align*}
We report held-out negative log predictive density,
\begin{align*}
  -\mathrm{lpd}
  =
  -\frac{1}{4n}
  \sum_{i=1}^n \sum_{t=13}^{16}
  \log \hat\pi_{it,x_{it}},
\end{align*}
so lower values are better.

\begin{figure}[t]
\centering
\includegraphics[width=0.72\textwidth]{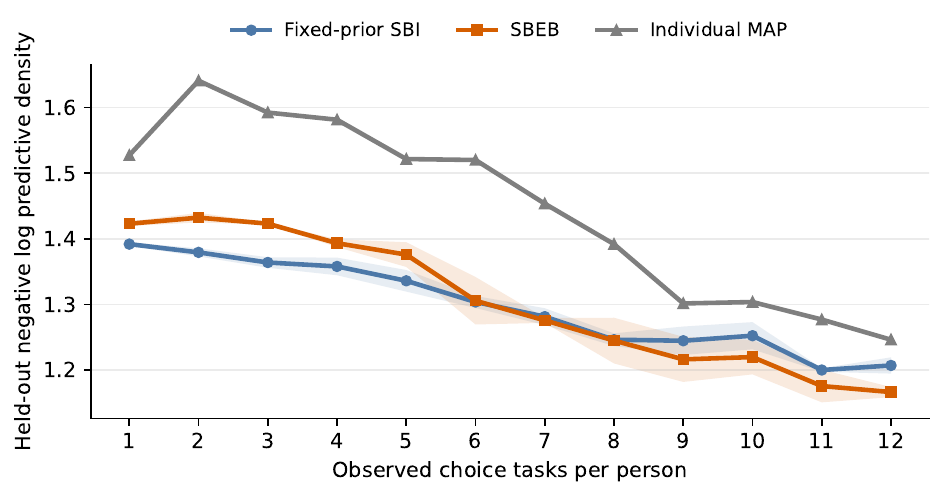}
\caption{Heterogeneous choice study.  Held-out negative log predictive density
on tasks 13--16 as a function of the number of observed training tasks per
person.  Each point averages three independent runs; bands show one standard
error.  Lower is better.}
\label{fig:choice-learning-curve}
\end{figure}

\Cref{fig:choice-learning-curve} shows the results. The individual fits are poor when few choices are observed.  Fixed-prior SBI is strongest when we observe few choices.  As more choices are observed, SBEB improves and becomes best when $R=9,\ldots,12$. The learned population prior helps once the per-person data contain enough signal to identify how the population should shrink.

\subsection{Measles outbreaks in the U.K.}
\label{subsec:measles-study}

The measles study uses biweekly reported measles cases for 40 U.K. cities
from 1944 to 1965, along with city populations and births.  These data are
available through the \texttt{spatPomp} ecosystem and are analyzed in
\citet{Dalziel:2016}.  The goal is to infer city-specific epidemic
parameters from a past segment of a city's time series and then reconstruct
future disease burden in that same city.

We use a stochastic epidemic simulator.  It is a TSIR-style model with seasonal transmission and noisy reported cases, following the tradition of mechanistic measles time-series models \citep{FinkenstadtGrenfell:2000,Bjornstad:2002} and partially observed Markov process simulators \citep{KingNguyenIonides:2016}.  The simulator encodes transmission, seasonality, births, and imperfect reporting.  The likelihood is intractable because it requires summing over latent epidemic trajectories, but simulation is straightforward.  \Cref{app:measles-details} gives the exact simulator equations and parameter transformations.

Let $x_i = x_{i,1:T}$ be the reported measles count for city $i$.  Let $a_i = a_{i,1:T}$ contain the known inputs to the simulator, including population, births, and calendar time.  The latent parameter $z_i$ contains city-specific epidemic parameters: baseline transmission, seasonal amplitude and phase, reporting rate, importation, nonlinear mixing, observation overdispersion, and the initial susceptible fraction.  The simulator draws $x_i$ from $\rmp(x_i \g z_i, a_i)$.

Our algorithms fit an inference network to approximate the posterior distribution of the latent simulator parameters. For a training length $R$, the inference network receives summaries $s_i^{(R)}$ computed from the $R$ periods before the held-out period and from the known inputs $a_i$.  These summaries include log case counts, log incidence, the fraction of zero counts, the maximum log case count, autocorrelations, average birth rate, and seasonal Fourier components.  The network maps $s_i^{(R)}$ to a diagonal Gaussian approximation $\rmq(z_i; s_i^{(R)}, \varphi)$ over an eight-dimensional unconstrained parameter vector.

For evaluation, we always hold out biweeks $200,\ldots,300$.  For city
$i$, the observed held-out mean log incidence is
\begin{align*}
  m_i^{\mathrm{obs}}
  =
  \frac{1}{101}
  \sum_{t=200}^{300}
  \log\left(1 + \frac{10^5 x_{it}}{N_{it}}\right),
\end{align*}
where $N_{it}$ is the city population.  For posterior draws
$z_i^{(1)},\ldots,z_i^{(S)}$, we simulate held-out trajectories
$\tilde x_i^{(s)}$ under the same known inputs $a_i$ and compute
$\hat m_i$ by replacing $x_{it}$ with the simulated counts and averaging
over posterior draws.  We report
\begin{align*}
  \frac{1}{n}\sum_{i=1}^n |\hat m_i - m_i^{\mathrm{obs}}|.
\end{align*}

\begin{figure}[t]
\centering
\includegraphics[width=0.72\textwidth]{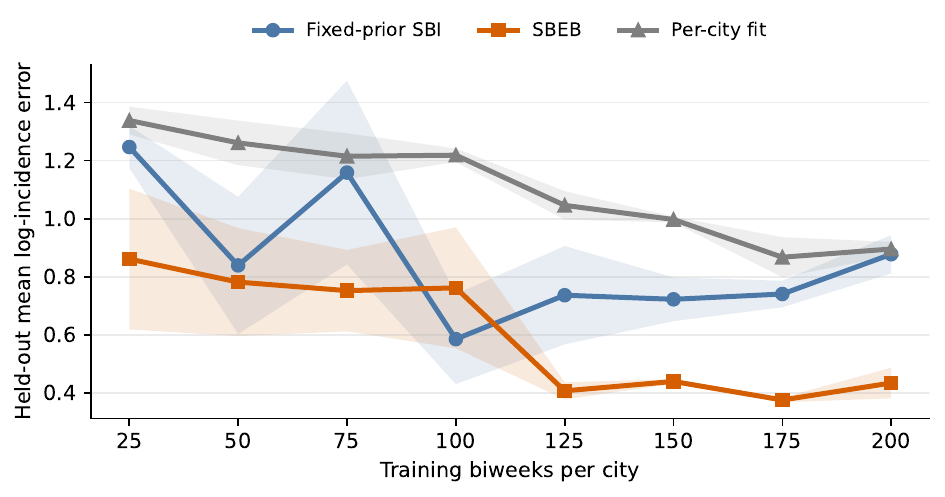}
\caption{Measles outbreak study.  Held-out mean log-incidence absolute error as
a function of the number of observed training biweeks per city.  Training uses
the $R$ biweekly observations immediately before observation 200.  Evaluation
is always on observations 200 through 300.  Each point averages three
independent runs; bands show one standard error.  Lower is better.}
\label{fig:measles-learning-curve}
\end{figure}

Again, we compare fixed-prior SBI, SBEB, and individual per-city
models.  The fixed prior is a broad Gaussian prior on the
unconstrained simulator parameters, centered on plausible epidemic
values before observing the dataset; \Cref{app:measles-details} gives
the values.  SBEB starts from the same prior and learns an empirical
Bayes prior.  The individual model fits each city separately.

\Cref{fig:measles-learning-curve} shows the results.  SBEB has the lowest average error for most training lengths, with the clearest gains at longer histories.  The learned EB prior helps the city-specific posterior produce simulated future epidemics with more accurate average incidence than either the fixed prior or the per-city calibration baseline.


\section{Discussion}

Simulation-based empirical Bayes (SBEB) brings EB to models whose
likelihoods are intractable.  Classical EB estimates prior structure
from many related observations, but usually assumes a tractable
likelihood.  SBI handles intractable likelihoods by training an
amortized posterior from simulator samples, but usually assumes a
fixed prior.  SBEB combines these ideas.

SBEB is useful in settings where empirical Bayes is useful, but where the likelihood is available only through simulation.  It expands the scope of EB from models with tractable likelihoods to models defined by scientific simulators.

Compared to traditional methods for EB and SBI, the price of SBEB is
more computation and more modeling choices.  SBEB is more expensive
than SBI with a fixed prior because it runs in rounds.  It is more
expensive than EB with a simple likelihood because it requires
simulation-based inference.

It also inherits practical issues from both fields.  SBI requires an amortized family expressive enough to perform inference; see \citet{MargossianBlei:2023} for a discussion of when amortized inference can work well.  And SBEB is built on the premise that the simulator gives an accurate conditional distribution of datasets.  The practitioner remains responsible for developing and checking both the high-fidelity simulator and a good amortized family.


\section*{Acknowledgements}

The authors thank Sebastian Wagner-Carena for useful discussions and feedback.


\bibliographystyle{apalike}
\bibliography{sbi_eb.bib}
\newpage
\appendix
\renewcommand{\thesection}{\Alph{section}}
\titleformat{\section}
  {\normalfont\large\bfseries}
  {Appendix~\thesection}
  {1em}
  {}
\section{Proofs}
\label{app:proofs}

\subsection*{Proof of \Cref{eq:pop_prior}}

Our goal is to set the prior $\rmp(z)$ so that the marginal $\rmp(x)$
equals the population distribution $\rmp^\star(x)$.

To fit the prior, it suffices to solve
\begin{align}
  \label{eq:eb_optimization}
\rmp(z) = \arg \max_{\rmp(z)} \EE{\rmp^\star(x)}{\log \rmp(x)}.
\end{align}
Recall that the marginal $\rmp(x)$ is a function of the prior $\rmp(z)$ through
\Cref{eq:marginal}.
\Cref{eq:eb_optimization} solves the problem
because
it maximizes the expected logarithmic score of the induced marginal $p(x)$,
and the logarithmic scoring rule is strictly proper~\citep{Gneiting:2007}.
Thus,  under realizability,
this objective is maximized when $\rmp(x) = \rmp^\star(x)$.
If the simulator is  prior-identifiable, the maximizing prior is unique.

We now solve the optimization problem.
Since $p(z)$ is a probability density, the optimization is constrained: $\int \rmp(z) \dd z = 1$.
We form the Lagrangian and expand the marginal,
\begin{align*}
  \cL(\rmp(z))
&= \EE{\rmp^\star(x)}{\log \rmp(x)}
    - \lambda\left(\int \rmp(z) \dd z - 1\right) \\
&= \EE{\rmp^\star(x)}{\log \int \rmp(z) \rmp(x \g z) \dd z}
    - \lambda \left(\int \rmp(z) \dd z - 1\right).
\end{align*}
Now we take the derivative with respect to $\rmp(z)$,
\begin{align}
  \frac{\partial{\cL}}{\partial \rmp(z)}
  &=
\EE{\rmp^\star(x)}
    {\frac{\rmp(x \g z)}{\int \rmp(z') \rmp(x \g z') \dd z'}}
    - \lambda \nonumber \\
&= \EE{\rmp^\star(x)}
    {\frac{\rmp(z \g x) \rmp(x)}{\rmp(x) \rmp(z)}}
    - \lambda \label{eq:line_2}\\
    &= \frac{\EE{\rmp^\star(x)}{\rmp(z \g x)}}{\rmp(z)} - \lambda.
    \label{eq:line_3}
\end{align}
\Cref{eq:line_2} replaces the integral with $\rmp(x)$ and uses a
(reverse) Bayes rule, \[\rmp(x \g z) = \rmp(z \g x) \rmp(x) / \rmp(z).\]
\Cref{eq:line_3} implies the optimal prior is proportional to the
population-expectation of the posterior,
\begin{align*}
\rmp(z) \propto \EE{\rmp^\star(x)}{\rmp(z \g x)}.
\end{align*}
Normalizing both sides gives \Cref{eq:pop_prior}.

\subsection*{Proof of \Cref{prop:monotonicity}}
Here, we prove the monotonicity of the population EB update.
First note that the objective in \Cref{eq:lowerbound}
is a lower bound on the log marginal likelihood via Jensen's inequality:
for any prior~$p(z)$, its induced marginal~$p(x)$, and  $\rmq(z; x)$,
\begin{equation}
\label{eq:jensen}
\EE{\rmq(z; x)}{\log \frac{\rmp(z)\rmp(x\g z)}{\rmq(z; x)}} \leq \log \rmp(x),
\end{equation}
with equality if and only if
 \begin{equation}
 \rmq(z; x) = \rmp(z\g x).
 \end{equation}
 At iteration $t$ of the idealized update,
 the E-step recovers the posterior under the current prior,
 i.e.,
 $q(z; x) = p_t(z \g x)$.
Then \Cref{eq:jensen} is tight, and after taking population expectations,
we obtain
\begin{align}
\EE{\rmp^\star(x)}{\log \rmp_t(x)} \,&= \, \ell(\rmp_t(z), \rmp_t(z\g x)).
\end{align}
Holding $p_t(z\g x)$ fixed, the M-step chooses $p_{t+1}$ to maximize the objective $\ell$ over priors,
which implies that
\begin{align*}
\ell(\rmp_t(z), \rmp_t(z\g x)) \leq \, \ell(\rmp_{t+1}(z),\rmp_t(z\g x)). 
\end{align*}
Another application of Jensen's inequality implies monotonicity of the update:
\begin{align*}
\EE{\rmp^\star(x)}{\log \rmp_t(x)}
\leq \, \ell(\rmp_{t+1}(z),\rmp_t(z\g x))
\leq\,    \EE{\rmp^\star(x)}{\log \rmp_{t+1}(x)}.
\end{align*}
Finally, dividing the marginal density on both sides by $p^*(x)$ obtains the equivalent KL
divergence inequality.

\subsection*{Proof of \Cref{prop:dlemma}}

Here, we show that the population EB update decreases the KL divergence to the population prior by
at least the KL divergence of the marginals.
We begin by rewriting
the update in  \Cref{eq:oracle_prior_update} as
\begin{align*}
\rmp_{t+1}(z) &= \int \rmp_t(z\g x) \rmp^\star(x) \, \dd x \\
&= \, \int \frac{\rmp_t(z)\rmp(x\g z)}{\rmp_t(x)}\rmp^\star(x)\, \dd x \\
&= \rmp_t(z) \EE{\rmp(x\g z)}{\frac{\rmp^\star(x)}{\rmp_t(x)}}.
\end{align*}
Therefore,
\begin{align}
\label{eq:priorratio}
\frac{\rmp_{t+1}(z)}{\rmp_{t}(z)} &= \EE{\rmp(x\g z)}{\frac{\rmp^\star(x)}{\rmp_t(x)}}.
\end{align}

Now we consider the decrease in the prior KL:
\begin{align}
\kl{\rmp^\star(z) \, \Vert \, \rmp_{t}(z)} - \kl{\rmp^\star(z) \, \Vert \, \rmp_{t+1}(z)}  &= \,
    \EE{\rmp^\star(z)}{\log\frac{\rmp_{t+1}(z)}{\rmp_t(z)}} \nonumber \\
\label{eq:ratio2}
    &= \EE{\rmp^\star(z)}{\log \, \EE{\rmp(x\g z)}{\frac{\rmp^\star(x)}{\rmp_t(x)}}} \\
\label{eq:jensen2}
&\geq \, \EE{\rmp^\star(z)}{\EE{\rmp(x\g z)}{\log\frac{\rmp^\star(x)}{\rmp_t(x)}}} \\
&= \, \EE{\rmp^\star(x)}{\log\frac{\rmp^\star(x)}{\rmp_t(x)}} \nonumber \\
&= \, \kl{\rmp^\star(x) \, \Vert \, \rmp_{t}(x)}, \nonumber
\end{align}
where \Cref{eq:ratio2} follows from
applying \Cref{eq:priorratio} and
\Cref{eq:jensen2} follows from Jensen's inequality.

\subsection*{Proof of \Cref{cor:marginal-convergence}}

Here we show that the induced marginal converges to the population marginal in KL divergence.
We define the quantities
\begin{align*}
a_t = \kl{\rmp^\star(x) \, \Vert \, \rmp_t(x)}
\quad \textrm{and} \quad
D_t = \kl{\rmp^\star(z) \, \Vert \, \rmp_t(z)}.
\end{align*}
By \Cref{prop:monotonicity}, the marginal KL sequence $\{a_t\}$ is nonincreasing.
By
\Cref{prop:dlemma}, each marginal KL satisfies $a_t \leq D_t - D_{t+1}$.
Summing this inequality
gives
\begin{align*}
\sum_{t=0}^{T} a_t
\leq
D_0 - D_{T+1}
\leq
D_0.
\end{align*}
By assumption, $D_0 < \infty$, and so the nonnegative sequence $\{a_t\}$ has a finite sum.
Since it is also nonincreasing, its limit must be zero.

\subsection*{Proof of \Cref{thm:convergence}}
Our goal is to prove the geometric convergence of the KL divergence of prior update,
and then to use this result with the data processing inequality to derive a bound on the marginal KL values.

First, by the descent lemma in \Cref{prop:dlemma},
for each update $t \geq 0$,
\begin{align*}
\kl{\rmp^\star(z) \, \Vert \, \rmp_{t+1}(z)} \, &\leq \, \kl{\rmp^\star(z) \, \Vert \, \rmp_{t}(z)}
    \,- \, \kl{\rmp^\star(x) \, \Vert \, \rmp_{t}(x)}.
\end{align*}
By $\lambda$-strong identifiability of the prior (\Cref{def:prior}),
\begin{align*}
    \kl{\rmp^\star(z) \, \Vert \, \rmp_{t}(z)} \, &\leq \, \tfrac{1}{\lambda} \kl{\rmp^\star(x) \,
    \Vert \, \rmp_{t}(x)}.
\end{align*}
Substituting this inequality into the descent inequality, we obtain
\begin{align*}
\kl{\rmp^\star(z) \, \Vert \, \rmp_{t+1}(z)} \, &\leq \,
    \kl{\rmp^\star(z) \, \Vert \, \rmp_{t}(z)} \,- \,
    \lambda \kl{\rmp^\star(z) \, \Vert \, \rmp_{t}(z)}
    \\
& = \, (1-\lambda) \, \kl{\rmp^\star(z) \, \Vert \, \rmp_{t}(z)}.
\end{align*}
Iterating this inequality proves the first claim.
The second claim follows directly from
applying the data processing inequality, i.e.,
\begin{align*}
\kl{\rmp^\star(x) \, \Vert \, \rmp_t(x)} \leq \kl{\rmp^\star(z) \, \Vert \, \rmp_t(z)},
\end{align*}
to the first claim.

\section{Simulation-study details}
\label{app:simulation-details}

This appendix gives the mathematical definitions of the five simulators used in
\Cref{subsec:simulation-studies}.  In each case, the inference algorithms use
only simulated pairs $(z,x)$; they do not evaluate the likelihood.

\paragraph{Linear Gaussian model.}
The latent variable is $z \in \mathbb R^3$.  A fixed matrix
$A \in \mathbb R^{10\times 3}$ maps $z$ to a ten-dimensional observation,
\begin{align}
  x = A z + \epsilon,
  \qquad
  \epsilon \sim \mathcal N(0, 0.3^2 I).
\end{align}
The true prior is a standard Gaussian.  The fixed-prior SBI baseline uses a
shifted and rescaled Gaussian prior.

\paragraph{Nonlinear static model.}
The latent variable is $z \in \mathbb R^3$ and the observation is
$x\in\mathbb R^5$,
\begin{align}
  x_1 &= \sin(z_1) + 0.3 z_2 + \epsilon_1, \\
  x_2 &= \cos(z_2) + 0.2 z_3 + \epsilon_2, \\
  x_3 &= \tanh(z_3) + 0.1 z_1 + \epsilon_3, \\
  x_4 &= z_1 z_2 + \epsilon_4, \\
  x_5 &= z_2 z_3 + \epsilon_5,
\end{align}
with independent Gaussian noise.  The true and fixed priors follow the same
pattern as in the linear Gaussian model.

\paragraph{Damped oscillator.}
The latent parameter is $z=(\omega,\gamma)$, where $\omega$ is the angular
frequency and $\gamma$ is damping.  The state $(q,v)$ evolves according to
\begin{align}
  \dot q &= v, \\
  \dot v &= -\omega^2 q - \gamma v,
\end{align}
with initial condition $q(0)=1$ and $v(0)=0$.  The observation is the noisy
trajectory $x=(q(t_1),\ldots,q(t_T)) + \epsilon$ on a fixed time grid.  The true
prior places $\omega\sim\mathrm{Unif}(0.5,1.5)$ and
$\gamma\sim\mathrm{Unif}(0.05,0.3)$; the fixed-prior baseline shifts these
ranges upward.

\paragraph{Predator-prey dynamics.}
The latent parameter is $z=(\alpha,\beta,\delta,\gamma)$ in the
Lotka-Volterra system,
\begin{align}
  \dot y_1 &= \alpha y_1 - \beta y_1 y_2, \\
  \dot y_2 &= \delta y_1 y_2 - \gamma y_2,
\end{align}
with initial condition $y_1(0)=1$ and $y_2(0)=0.5$.  We integrate the system on
a fixed grid and observe the noisy trajectories of both species.  The true prior
places $\alpha,\gamma\sim\mathrm{Unif}(0.8,1.2)$ and
$\beta,\delta\sim\mathrm{Unif}(0.05,0.10)$; the fixed-prior baseline shifts
these ranges.

\paragraph{Wright--Fisher evolutionary dynamics.}
The simulator tracks allele frequencies in two populations across many
independent loci.  The unconstrained latent variable is four-dimensional and is
transformed to effective population sizes, migration, and mutation parameters,
\begin{align}
  N_1 &= \exp(z_1) + 50, &
  N_2 &= \exp(z_2) + 50, \\
  m &= 0.2\,\sigma(z_3), &
  \mu &= 0.02\,\sigma(z_4),
\end{align}
where $\sigma(\cdot)$ is the logistic function.  At each generation and locus,
the next allele frequencies are binomial draws whose probabilities include
migration between populations and mutation.  After the final generation, the
observation is a summary of the allele frequencies,
\begin{align}
  x = (\bar p_1,\bar p_2,\mathrm{Var}[p_1],\mathrm{Var}[p_2],F_{ST}),
  \qquad
  F_{ST}
  =
  \frac{\mathbb E[(p_1-p_2)^2]}{\mathbb E[p_1(1-p_1)]}.
\end{align}
The true prior generates moderate population sizes and migration/mutation
rates; the fixed-prior baseline shifts population sizes and the
migration/mutation parameters upward.

\section{Measles simulator details}
\label{app:measles-details}

The measles study uses a stochastic TSIR-style simulator.  For city $i$, let
$N_{it}$ be population and $b_{it}$ births at biweek $t$.  The simulator tracks
susceptible individuals $S_t$, latent infections $I_t$, and reported cases
$Y_t$.  Given parameters
\[
z_i = (\beta_0, a, \phi, \rho, \iota, \alpha, \kappa, s_0),
\]
the seasonal transmission rate is
\begin{align}
  \beta_t
  =
  \beta_0 \exp\left\{
  a \cos\left(2\pi(t \bmod 26)/26 + \phi\right)
  \right\}.
\end{align}
The latent force of infection is
\begin{align}
  \lambda_t
  =
  \frac{\beta_t S_t (I_t + \iota)^\alpha}{N_{it}}.
\end{align}
The simulator draws latent infections and reported cases as
\begin{align}
  I_{t+1} &\sim \mathrm{Poisson}(\lambda_t), \\
  Y_t &\sim \mathrm{NegBinom}(\rho I_{t+1}, \kappa),
\end{align}
where the negative binomial is parameterized by its mean and overdispersion.
The susceptible population evolves as
\begin{align}
  S_{t+1} = \max\{S_t + b_{it} - I_{t+1}, 1\},
  \qquad
  S_0 = s_0 N_{i0}.
\end{align}

The inference network works on an unconstrained vector
$\eta\in\mathbb R^8$.  Before simulation, the code transforms it to epidemic
parameters:
\begin{align}
  \beta_0 &= \exp(\eta_1), &
  a &= 1.5\,\sigma(\eta_2), &
  \phi &= \pi \tanh(\eta_3), \\
  \rho &= 0.02 + 0.96\,\sigma(\eta_4), &
  \iota &= \exp(\eta_5), &
  \alpha &= 0.85 + 0.25\,\sigma(\eta_6), \\
  \kappa &= 2 + 80\,\sigma(\eta_7), &
  s_0 &= 0.01 + 0.20\,\sigma(\eta_8),
\end{align}
where $\sigma(u) = 1/(1+\exp(-u))$.

The fixed prior for SBI is a Gaussian on the unconstrained scale,
\[
  \eta \sim \mathcal N(m, I).
\]
We choose $m$ by first choosing plausible epidemic-scale values,
\[
  (\beta_0, a, \phi, \rho, \iota, \alpha, \kappa, s_0)
  =
  (2.5, 0.7, 0, 0.10, 0.05, 0.97, 20, 0.04),
\]
and then applying the inverse of the transformations above.  This gives
\[
  m =
  (0.916, -0.134, 0, -2.398, -2.996, -0.080, -1.237, -1.735).
\]


\end{document}